# DeStress: Deep Learning for Unsupervised Identification of Mental Stress in Firefighters from Heart-rate Variability (HRV) Data

Ali Oskooei[1], Sophie Mai Chau[1], Jonas Weiss[1], Arvind Sridhar[1], María Rodríguez Martínez[1] and Bruno Michel[1]

**Abstract**—In this work we perform a study of various unsupervised methods to identify mental stress in firefighter trainees based on unlabeled heart rate variability data. We collect RR interval time series data from nearly 100 firefighter trainees that participated in a drill. We explore and compare three methods in order to perform unsupervised stress detection: 1) traditional K-Means clustering with engineered time and frequency domain features 2) convolutional autoencoders and 3) long short-term memory (LSTM) autoencoders, both trained on the raw RRI measurements combined with DBSCAN clustering and K-Nearest-Neighbors classification. We demonstrate that K-Means combined with engineered features is unable to capture meaningful structure within the data. On the other hand, convolutional and LSTM autoencoders tend to extract varying structure from the data pointing to different clusters with different sizes of clusters. We attempt at identifying the true stressed and normal clusters using the HRV markers of mental stress reported in the literature. We demonstrate that the clusters produced by the convolutional autoencoders consistently and successfully stratify stressed versus normal samples, as validated by several established physiological stress markers such as RMSSD, Max-HR, Mean-HR and LF-HF ratio.

## 1 Introduction

Stress is the difficulty of an organism to maintain its homeostasis, often induced by external stimuli that cause mental or physical imbalance [1]. It is known that when an individual is exposed to a stressor, the autonomic nervous (ANS) system is triggered resulting in the suppression of the parasympathetic nervous system and the activation of sympathetic nervous system [2]. This reaction which is known as the fight-or-flight response can involve physiological manifestations such as: vasoconstriction of blood vessels, increased blood pressure, increased muscle tension and a change in heart rate (HR) and heart rate variability (HRV) [3], [4]. Among these, HRV has become a standard metric for the assessment of the state of body and mind, with multiple markers derived from HRV being routinely used for identifying mental stress or lack thereof. HRV is a time series of the variation of the heart rate over time and is determined by calculating the difference in time between two consecutive occurrences of QRS-complexes, also known as the RR interval (RRI) [5], [6]. An optimal HRV points to healthy physiological function, adaptability and resilience. Increased HRV (beyond normal) may point to a disease or abnormality. Reduced HRV, on the other hand, points to an impaired regulatory capacity and is known to be a sign of stress, anxiety and a number of other health problems [4], [7], [8].

Identifying stress has been the focus of much research as an increasing body of evidence suggests a rising prevalence of stress-related health conditions associated with the stressful contemporary lifestyle [9]. A significant portion of the contemporary stress is due to the occupational stress. Occupational stress can not only result in chronic health conditions such as heart disease [10] but can also have more immediate catastrophic effects such as accidents, injury and even death [11], [12]. Firefighters and smoke divers, in particular, are susceptible to acute stress due to the sensitive nature of their work. It is imperative that mental stress in firefighters is monitored to prevent injury to personnel or the public [13], [14]. In this work, our objective is to leverage HRV data and unsupervised machine learning methods, in order to detect mental stress in firefighters.

With the rise of modern machine learning and deep learning methods, these methods have been applied in the study of heart rate variability. Machine learning and deep learning methods have previously been used with HRV and electrocardiography (ECG) data for various applications such as: fatigue and stress detection [15]–[20], student stress prediction [21], congestive heart failure detection [17], [22], cardiac arrhythmia classification [23], [24]. The vast majority of prior arts, however, are supervised or

- *Ali Oskooei is with IBM Research – Zurich, Säumerstrasse 4, 8803 Rüschlikon, Switzerland. E-mail: osk@zurich.ibm.com.*
- *Sophie Mai Chau is with IBM Denmark –KONGEVEJEN 495B HOLTE, 2840, Denmark. E-mail: sophie.mai.chau@ibm.com.*
- *Jonas Weiss is with IBM Research – Zurich, Säumerstrasse 4, 8803 Rüschlikon, Switzerland. E-mail: jwe@zurich.ibm.com.*
- *Arvind Sridhar is with IBM Research – Zurich, Säumerstrasse 4, 8803 Rüschlikon, Switzerland. E-mail: rvi@zurich.ibm.com.*
- *María Rodríguez Martínez is with IBM Research – Zurich, Säumerstrasse 4, 8803 Rüschlikon, Switzerland. E-mail: mrm@zurich.ibm.com.*
- *Bruno Michel is with IBM Research – Zurich, Säumerstrasse 4, 8803 Rüschlikon, Switzerland. E-mail: bmi@zurich.ibm.com.*

based on labeled public datasets as opposed to unlabeled real-world data. There have been few previous attempts at unsupervised detection of mental stress, from ECG data, using for instance, self organizing maps (SOMs) [25] or traditional clustering algorithms [26]. To the best of our knowledge, deep learning based unsupervised detection of mental stress in firefighters using short-term HRV data has not been studied before. In this paper, we propose an unsupervised approach using autoencoders and density-based clustering, combined with prior knowledge, in order to cluster and label raw HRV data collected from firefighters.

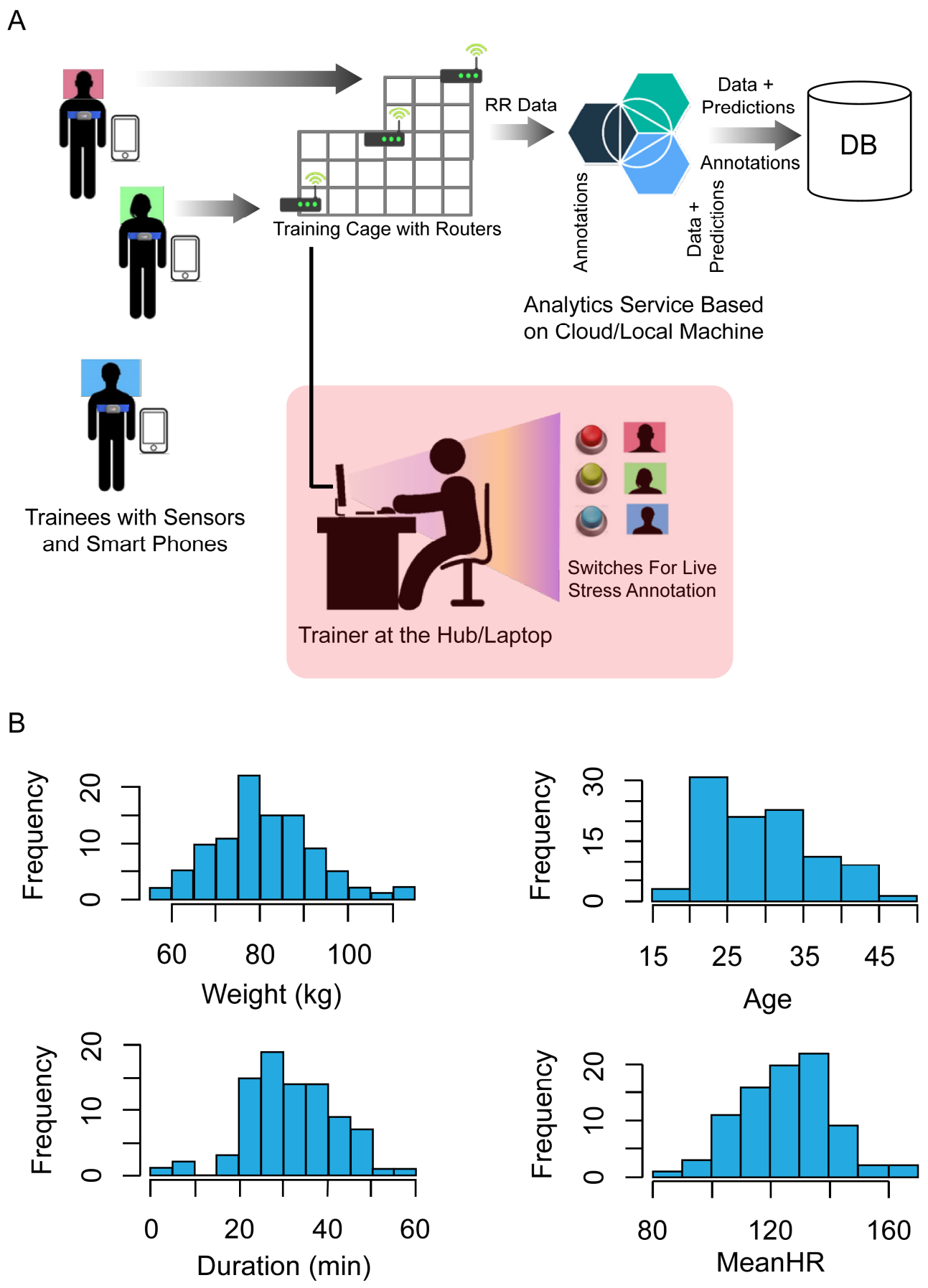

Fig. 1. A) Schematic illustration of data collection and expert labeling of stress in firefighter trainees. Each trainee wears a chest strap sensor and carries a smart phone that collects and transfers the HRV data to be analyzed, predicted or labeled. The expert labeling process, however was proven difficult and resulted in poor labeling of the data and the need for unsupervised classification of the collected HRV data. B) Statistics of the subject firefighter trainees: Weight in kilograms, Age, Duration of HRV data collected, Mean heart rate in beats per minute.

In this work, we collect RRI time series data from nearly 100 fire fighter trainees. We break down the collected time series data into 30 measurement windows prior to further processing or modeling. Successful calculation of stress markers from short-term HRV measurements (~10s) have been previously demonstrated [22], [27], [28]. We then use the HRV samples with both classical clustering methods as well as convolutional and long short-term memory (LSTM) autoencoders to find meaningful clusters in the context of mental stress detection.

## 2 Results and Discussion

### 2.1 Classical Feature Engineering and K-means

As a first attempt at unsupervised classification of HRV data, we employed K-means clustering on 18 engineered features (see supplementary table S1). The K-means clustering results for k = 2 are shown in figure 2C, plotted in two dimensions for mean heartrate (MeanHR) and root mean square of successive differences (RMSSD), two of the top reported biomarkers of stress in the literature[29], [30]. As shown in the figure, the identified clusters appear synthetic without a clear separation in the data. This is in part due to the high dimensionality of the data [31] as well as the intrinsic tendency of K-means clustering algorithm to cluster samples based on the Euclidean distance from the centroids of clusters, regardless of true separation within the data [32].

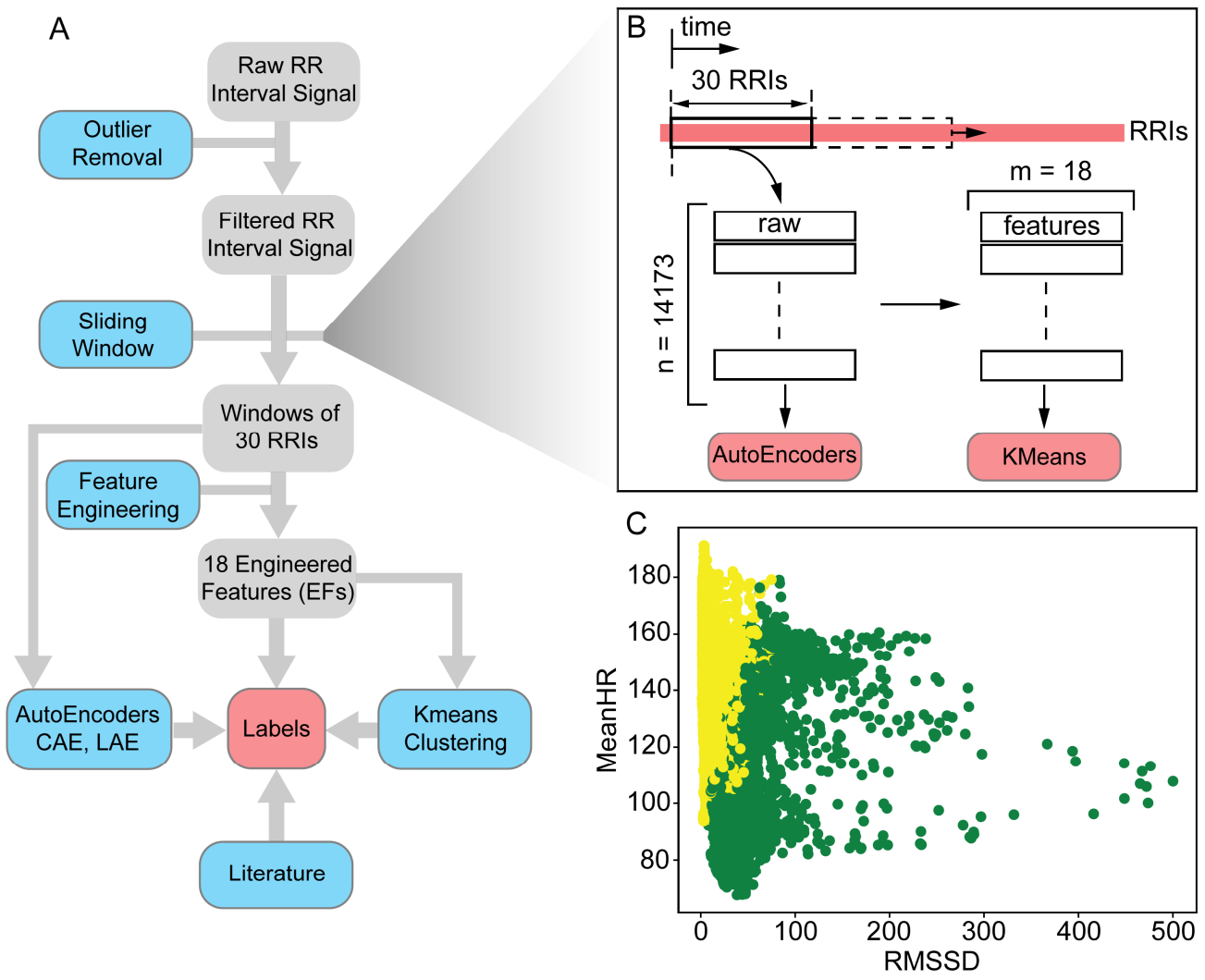

Fig. 2. A) The workflow for unsupervised classification of HRV data in this work B) sliding window transformation of the RRI into non-overlapping windows of 30 time steps and C) the results of K-means clustering with $k = 2$ on the samples using the 18 engineered features. As seen in the plot K-means does not produce distinct well-separated clusters within the data.

### 2.2 Convolutional and LSTM AutoEncoders

Having observed the inability of K-means clustering in finding meaningful structure in the data, we explored autoencoders to compress the data and find meaningful structures that we can leverage to identify mental stress. We trained and evaluated two autoencoder architectures: a convolutional autoencoder (CAE) and a LSTM autoencoder (LAE). The architectures of the two models are shown in figures 3A and 3B.

We adopted a 5-fold cross-validation scheme and trained five models for each autoencoder. The reconstruction errors across all fold for both CAE and LAE are shown in figure 3C. The LAE has slightly lower reconstruction error and results in a smoother reconstructed signal. A reconstructed validation sample using both CAE and LAE is presented in figure 3D.

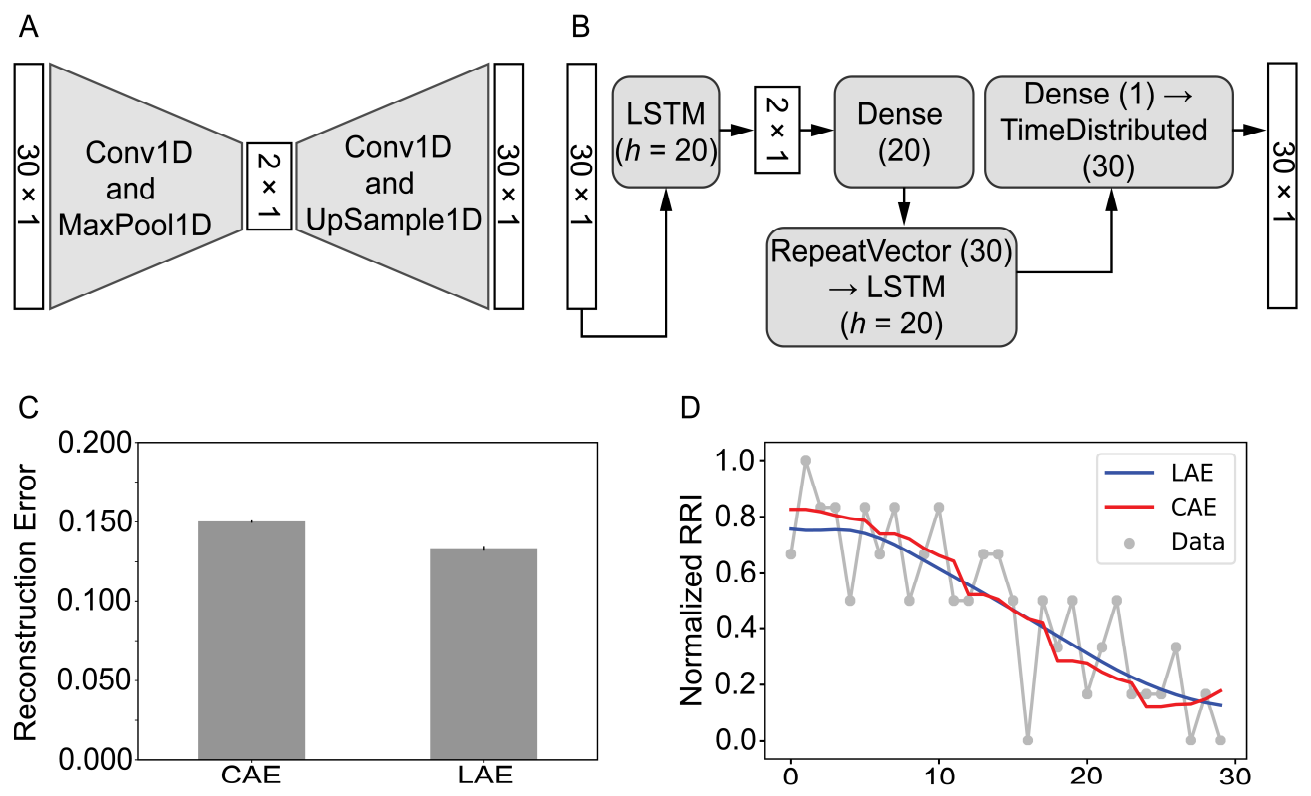

Fig. 3. The architecture and comparison of convolutional and LSTM autoencoders. A) The convolutional autoencoder with a 2D bottleneck consisting of 1D convolutions, maxpooling and upsampling layers. B) The LSTM autoencoder with a 2D bottleneck consisting of LSTM cells hidden dimension of 20. C) The reconstruction error (i.e., MAE) on the validation datasets across five folds for the CAE and the LAE. D) Reconstruction of a sample from the validation dataset by both the CAE and LAE.

Figures 4 and 5 show the training and validation results for the CAE and LAE models. On the top left corner of the figures, the validation error during training is plotted for all five models trained for five different folds. As demonstrated in the figures, the latent representation of the data encoded by either the CAE or the LAE exhibit separable clusters that could potentially point to a separation between the stressed and normal samples. The clusters generated by the CAE are well-separated and there is a large discrepancy between the sizes of the two clusters with nearly ~8% of the validation data in one cluster (y = 1) and the rest in the other cluster (y = 0). In addition, the CAE clusters are uniform in size across all five folds. Conversely, the LAE clusters are more balanced in size and the separation between the clusters is not significant. In addition, the sizes of LAE clusters are variable across different folds. The poor separation combined with the variable cluster size suggests that the LAE clusters do not represent a reproducible underlying structure within the data. We will investigate the meaningfulness of the clusters produced by each model shortly.

As described in section 4, in order to classify unseen data, we used a KNN classifier with the clusters and labels obtained from the validation results. The label prediction results on the unseen test samples are shown in figures 4 and 5 for the CAE and LAE models respectively. As shown in the figure 4, the test samples encoded by the CAE follow the same cluster pattern as the validation data with the majority of the samples in cluster "0" and the remaining samples in cluster "1".

Similarly, as seen in figure 5, the test samples encoded by the LAE follow the same pattern as the validation clusters. Having the test clusters in hand, the question was whether any of the CAE or LAE encoded clusters are meaningful in the context of mental stress detection. To address this question, we resorted to established HRV markers of mental stress and calculated and compared them across the two clusters for each of the CAE and LAE models. The objective was to determine whether samples in each cluster share specific characteristics pertaining mental stress that separate them from the samples in the other cluster.

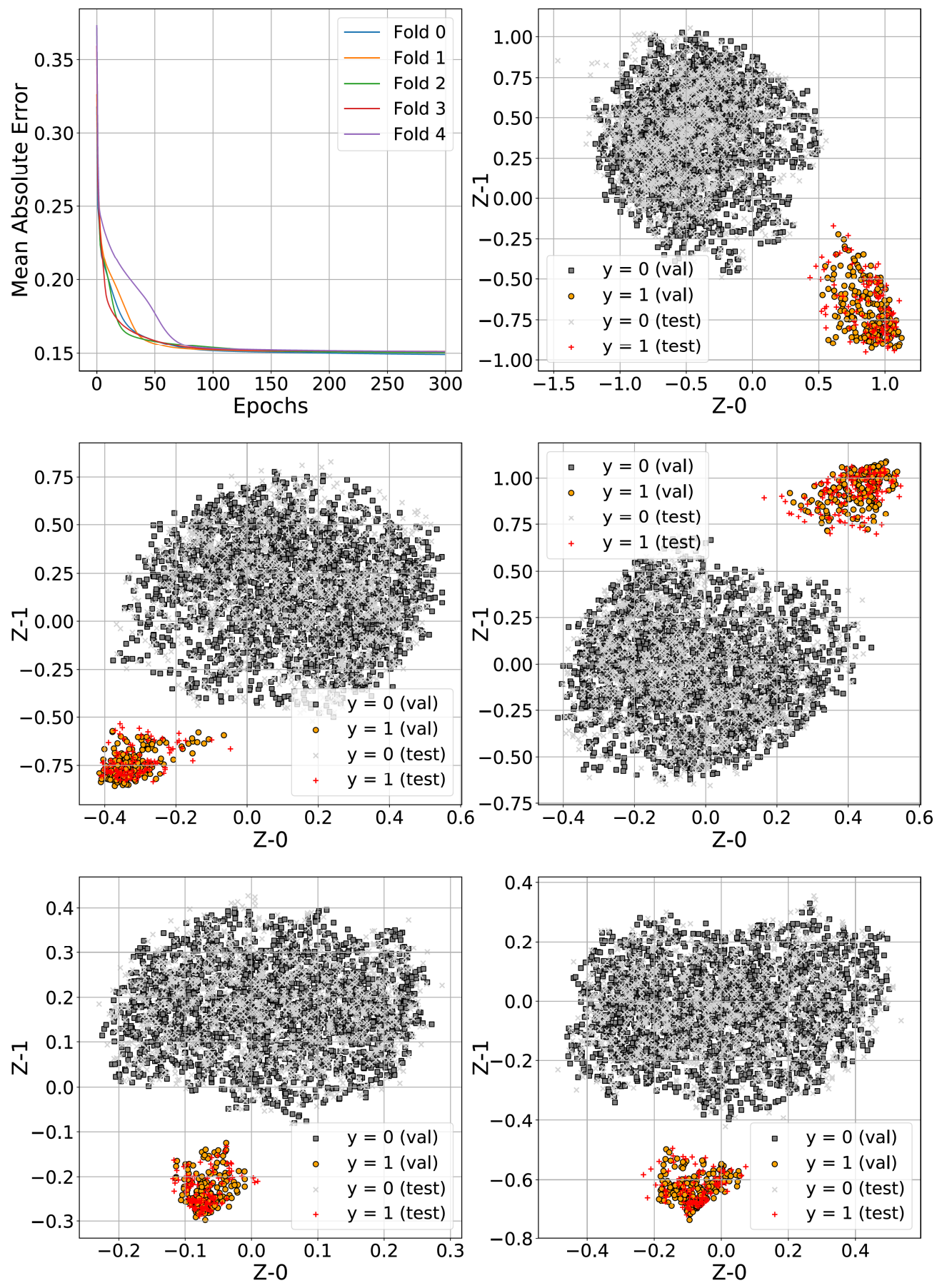

Fig. 4. Training and 5-fold cross-validation results for the CAE model as well as the predicted labels for the test dataset. The plot at the top left illustrates the validation error during the training of CAE in each fold. The scatter plots show the two clusters identified using DBSCAN clustering for the encoded validation data. We arbitrarily label the clusters “0” and “1”. Cluster “0” contains the majority of the data (~90%) while the rest belong to cluster “1”. The encoded test data and their KNN-predicted labels are superimposed with the validation clusters. As shown in the scatter plots, for all five folds, the encoded test data follow the same pattern as the validation data.

The mean RMSSD, a reported biomarker for mental stress [1], [9], [29], is compared across the two models and clusters in figure 6. As shown in figure 6, there is a significant (two-sided t-test, p-value = 2e-6) difference between the mean RMSSD for cluster "0" and cluster "1" of the test data encoded by the CAE. Cluster "0" (the grey bar) demonstrates a significantly lower RMSSD compared with cluster "1". The low RMSSD may signify low vagal tone and mental stress in cluster "0", as reported in the literature [33], [34]. Conversely, cluster "1" has high RMSSD, and as such a higher vagal tone indicating normal parasympathetic function [35].

On the other hand, the difference in mean RMSSD between the two LAE clusters are insignificant (two-sided t-test, p-value = 0.22), as shown in figure 6. The low discrepancy between the RMSSD of the LAE clusters further solidifies the hypothesis that the LAE-encoded clusters are not meaningful in the context of mental stress detection while the CAE-encoded clusters successfully stratify stressed versus normal samples. To further verify this hypothesis, we calculated and compared other HRV markers of stress across different clusters and models.

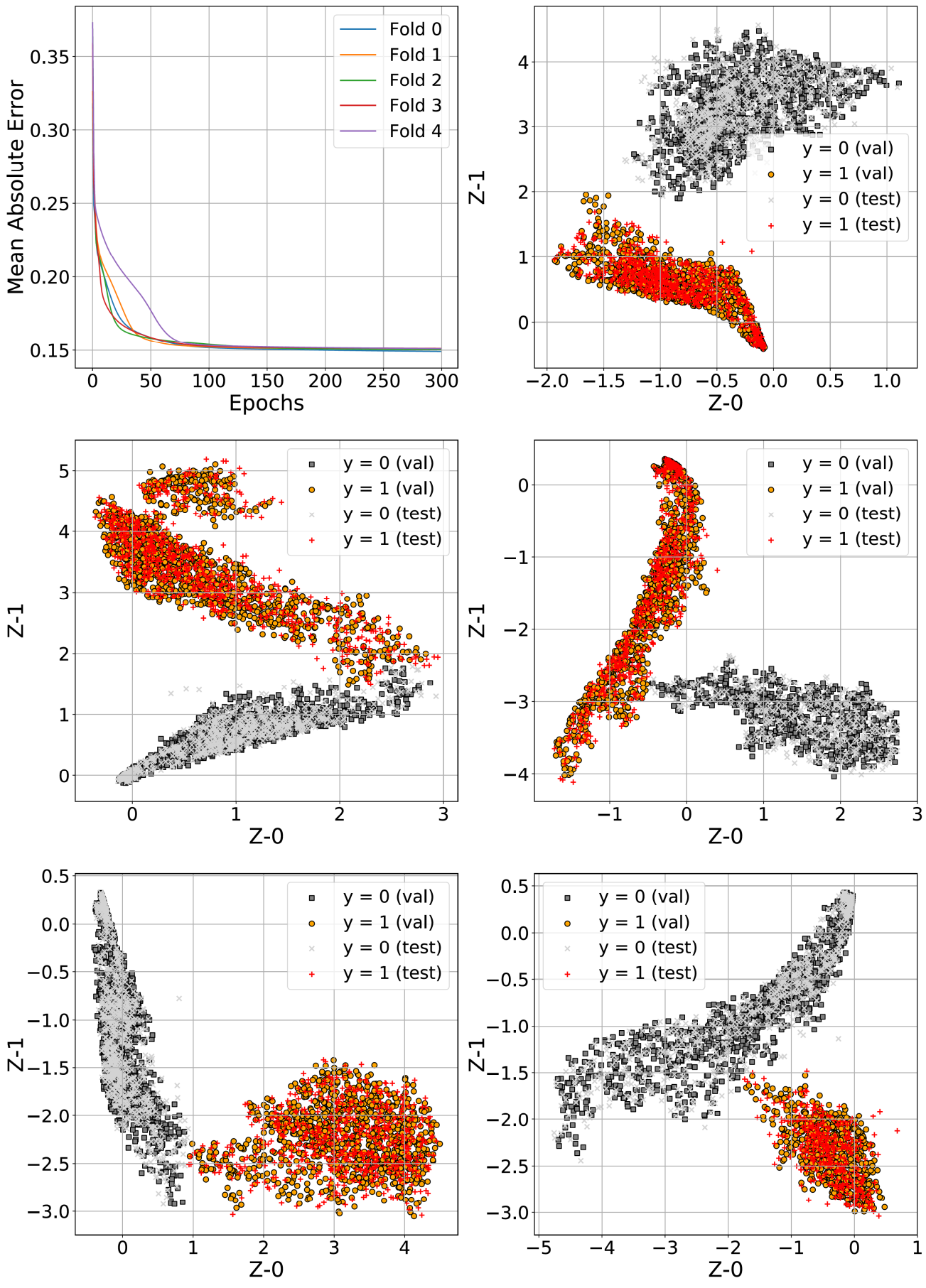

Fig. 5. Training and 5-fold cross-validation results for the LAE model as well as the predicted labels for the test dataset. The plot at the top left illustrates the validation error during the training of LAE in each fold. The scatter plots show the two clusters identified using DBSCAN clustering for the encoded validation data. We arbitrarily labeled the clusters “0” and “1”. In comparison with the CAE results, the LAE-encoded clusters exhibit poor separation. In addition, the cluster sizes are more balanced with significant variability across different folds. The encoded test data and their KNN-predicted labels are superimposed with the validation clusters. As shown in the scatter plots, for all five folds, the encoded test data follow the same pattern as the validation data.

The remaining barplots in figure 6, demonstrate a comparison between three other crucial HRV markers across the clusters given by the CAE and LAE model. Three features namely, maximum heart rate (Max-HR), mean RRI (Mean-RR) and low frequency (LF) to high frequency (HF) ratio (LF-HF Ratio) were selected based on their reported importance in detecting mental stress from HRV data [27], [29], [30]. As shown in figure 6, the CAE values show a statistically significant discrepancy between the two clusters for all three features: Max-HR (p-value = 7.2e-11), Mean-RR (p-value = 3.7e-8) and LF-HF Ratio (p-value = 4.8e-10). We observed that Max-HR is higher in CAE cluster "1" versus cluster "0" which is consistent with the reported correlation between Max-HR and RMSSD in the literature [3]. Interestingly, Mean-RR is lower in cluster "1" compared to cluster "0". This is while Mean-RR has been reported to decrease with the induction of mental stress compared with the resting state [2]. However, it should be noted that in our experiment, none of the subjects are in the resting state as they are all actively participating in a drill. LF-HF ratio is higher in cluster 0 than in cluster "1". Increased LF-HF ratio has been linked to mental stress in the literature [5]. In summary, two of the three HRV markers of stress suggest impaired vagal tone in cluster "0" and normal vagal tone for cluster "1". This is in agreement with the RMSSD results discussed earlier. Based on the results in figures 4 and 6, we postulate that the CAE cluster "1" corresponds to individuals that are physically stressed, as indicated by a high Max-HR and a slightly lower Mean-RR, but mentally relaxed, as indicated by a much higher RMSSD and a lower LF-HF ratio. CAE cluster "0", on the other hand, corresponds to samples experiencing both physical (i.e. high Max-HR and low Mean-RR) and mental stress (low RMSSD and high LF-HF Ratio).

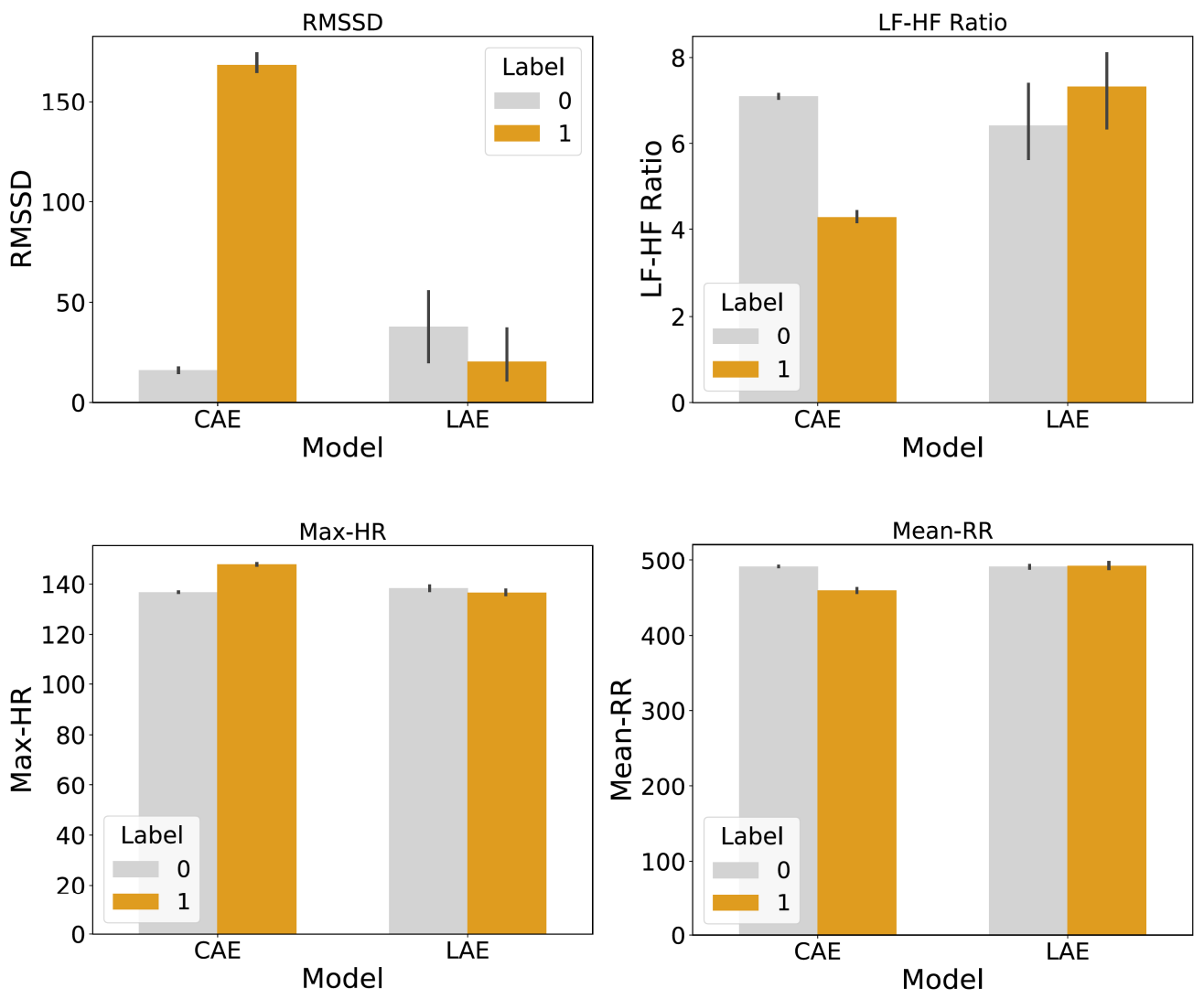

Fig. 6. Comparison of four HRV markers of stress across the clusters given by the CAE and LAE models. The CAE clusters show significant discrepancy in the HRV markers between the two clusters while the LAE clusters exhibit insignificant difference in the four HRV markers. These results further confirm our hypothesis that the CAE clusters are reproducible and relevant in the context of mental stress detection while the LAE clusters fail to stratify the samples in a meaningful manner.

For LAE clusters, on the other hand, HRV markers of stress do not exhibit significant discrepancy between the clusters: Max-HR (p-value = 0.05), Mean-RR (p-value = 0.66) and LF-HF Ratio (p-value = 0.22). As shown in figure 6, the barplots corresponding to the LAE clusters do not demonstrate a meaningful stratification of test samples based on the evaluated markers of mental stress. The results of figure 6 for the LAE clusters, are in line with our earlier observations and further confirm our hypothesis that LAE clusters do not appear meaningful in the context of mental stress detection.

## 2.3 Discussion

We studied various methods for unsupervised classification of HRV data collected from firefighter trainees. The objective was to detect mental stress from the HRV data. We started with time and frequency domain HRV features combined with traditional K-means clustering. The K-means clusters, however, appeared arbitrary and did not produce well-separated clusters within the data. To tackle this problem and bring out any existing hidden patterns within the data, we explored autoencoders (AEs). We incorporated two different neural net architectures namely the convolutional and LSTM neural nets and built two different autoencoders that were trained and evaluated separately. The LSTM autoencoder, LAE, achieved slightly lower reconstruction error than the convolutional autoencoder, CAE. We utilized the trained AEs to encode the validation samples into a 2D latent space in which the samples could be clustered in two separate clusters using the DBSCAN clustering algorithm. Each of the produced clusters were given an arbitrary identifier label ("0" and "1"). A K-Nearest-Neighbor classifier was fit to the validation data and the assigned labels for each validation dataset (i.e. fold). The resulting KNN models were used to make predictions for unseen samples. The KNN models were then used to classify test samples that were encoded by either of the CAE and LAE models. We observed nearly perfect agreement between the cluster patterns in the validation and test datasets. The clusters formed by the CAE encoding differed in size while the LAE encoding resulted in more equally sized but less separated clusters.

Having observed two separate clusters in the data for each model, we set out to determine two points: 1) clusters produced by which AE are the true meaningful clusters in the context of mental stress detection and 2) which cluster (or label) corresponds to mental stress. To answer these questions, we took advantage of well-established HRV markers of mental stress within the literature. We selected and determined four HRV markers of mental stress reported in the literature namely: RMSSD, Max-HR, Mean-RR and LF-HF Ratio. We observed that, for the CAE-encoded clusters, the values of these four markers showed a significant discrepancy across the clusters. In addition, the marker values for one cluster (cluster "0") were predominantly associated with mental stress according to the literature. For instance, cluster "0" had low RMSSD and high LF-HF ratio which are both associated with impaired vagal tone [33]. The LAE-encoded clusters, on the other hand, demonstrated insignificant discrepancy in the four evaluated markers across the two clusters. Based on this analysis, it was evident that LAE clusters fail to stratify the data in a manner that is meaningful for the identification of mental stress.

# 3 Data Acquisition

The RR intervals were collected from ~100 firefighters and smoke divers who volunteered to participate in this study. This study was conducted in March 2018 at the Kantonale Zivilschutzausbildungszentrum (Cantonal Civil Emergency services training center) of the Canton Aargau at Eiken AG in Switzerland where about 150 firefighters from various villages and public/private organizations across the Canton of Aargau participated in a weeklong training exercise to be certified as Smoke Diving team leaders. Among the various training exercises they participated in was the "Hotpot", a darkened chamber with a 3D maze made of cages, which simulated a building structure on fire. The goal was for firefighters to enter the maze in groups of two or three and find their way using only the illumination from their phosphorescent helmets. Smoke, strobing lights and loud noises are introduced during the exercise to disorient the participants and simulate real-world conditions involving heightened stress. A team of examiners observes the progress of the group from the outside of the chamber using infrared cameras. Along their way, the firefighters must also report any objects they find (cans containing chemicals, inflammable items etc.) to the examiners via radio communication. The firefighters were evaluated on their time of transit, their ability to find all the objects placed in the maze beforehand and the quality of the collaboration/team work within the group. This is an exercise that puts the participants through a period (10-20 minutes) of intense stress, both physical and mental. The participants were given heart rate measuring chest belts (Polar H10) and the "communication hub" smartphones to carry before they entered the maze. Once they finish the exercise, the devices were returned and a quick oral feedback on their perception of the difficulty of the exercise was noted. The RR interval time series were extracted from the ECG records. No ground truth regarding the mental stress status of the firefighters was given for the collected RR intervals. The outliers in the data were replaced with their nearest normal neighbors (i.e. Winsorized). The cleaned time-series data were then divided into non-overlapping windows of size 30 and shuffled as shown in figure 2B. 10% of the data was randomly selected and held out as the unseen test dataset and the remaining 90% was split into training and validation datasets according to a 5-fold cross-validation scheme. To enhance optimization and convergence of AEs, each raw RRI window of size 30 was scaled between 0 and 1 using min-max scaling. Time and frequency domain features (see supplementary table S1) were extracted using hrv-analysis 1.0.3 python library.

# 4 Methods

## 4.1 K-means Clustering

The traditional K-means clustering algorithm [36] was used to cluster the data transformed into 18 engineered features. We employed the K-means clustering algorithm implemented in Scikit-Learn Python library [37] with $k = 2$ and all other parameters set as default. However, common distance metrics, such as the Euclidean distance used in K-means, are not useful in finding similarity in high-dimensional data [31]. As a result, we explored autoencoders (AEs) as way to compress the raw samples into a lower dimensional latent space (2D in our case), and then search for patterns or clusters within the compressed (or encoded) data. We discuss autoencoders in detail in the following subsection.

### 4.2 AutoEncoders

Autoencoders are neural nets that ingest a sample, $x$, and attempt to reconstruct the sample at the output. When the autoencoder involves a hidden layer, $h$, that is of lower dimension than $x$, it is called an undercomplete autoencoder. The idea is to encode the data into a lower dimensional, $h$, which contains the most salient features of the data. The learning process of an autoencoder involves minimizing a loss function, $J$:

$$J = L(g(f(x)) \ where \ h = f(x) \tag{1}$$

where $f$ is the encoder, $g$ is the decoder and $L$ is a loss function that penalizes $g(h)$ for being dissimilar to $x$ [38]. We explored both mean squared error (MSE) and mean absolute error (MAE) as the loss function, $L$, and found mean absolute error to offer better convergence and lower reconstruction error compared with MSE.

As shown in figure 3A, in our convolutional autoencoder (CAE), both the encoder and the decoder consisted of four 1-dimensional convolution layers (kernel size = 2) and two maxpooling layers for the encoder and two upsampling layers for the decoder. Relu activation was used in all convolutional layers. The total number of trainable weights for the CAE was 710.

The architecture of the LSTM autoencoder (LAE) is shown in figure 3B. The encoder consists of a LSTM layer with hidden dimension of 20 followed by a linear transformation to the 2D bottleneck. The decoder consists of a dense transformation of the 2D bottleneck to 20 dimensions followed by vector repetitions (30 times) and a LSTM layer followed by a dense layer that reconstructs the input sample. Elu activation [39] was used in both the encoder and the decoder of the LAE. The total number of trainable weights for the LAE was 5163 which is an order of magnitude higher than the CAE.

### 4.3 DBSCAN Clustering

We employed DBSCAN clustering algorithm [40] to identify clusters in the latent representation of HRV data given by the AEs. DBSCAN is a density-based algorithm that clusters densely-packed samples together while disregarding samples in low-density areas as outliers.

### 4.4 Training and Evaluation

The AEs were implemented using tensorflow 1.12.0 (tf.keras) deep learning library. Adam optimizer [41] with a learning rate of 1e-4 and a batch size of 64 was used to train the AEs. 5-fold cross validation scheme was used to train the models and tune the hyperparamters. Both CAE and LAE were trained for 300 epochs. CAE loss plateaued after nearly 150 epochs while LAE plateaued much later at about 290 epochs. Both models were trained using a virtual machine with a 12-core CPU and 24 GB of RAM.

## 5 CONCLUSIONS

We presented a new approach for unsupervised detection of mental stress from raw HRV data using autoencoders. We demonstrated that classical K-means clustering combined with time and frequency domain features was not suitable for identifying mental stress. We then explored two different architectures of autoencoders to encode the data and find underlying patterns that may enable us to detect mental stress in an unsupervised manner. We trained convolutional and LSTM autoencoders and demonstrated that despite being more powerful and producing lower reconstruction error, LSTM autoencoders failed to identify useful patterns within the data. On the other hand, the convolutional autoencoders with their much fewer trainable weights, produced clusters that were verifiably distinct and pointed to different levels of mental stress according to the reported markers of mental stress. Based on the results given by the convolutional autoencoder, more than 90% of the samples collected from firefighter trainees during a drill were mentally stressed while less than 10% had normal HRV. While our proposed approach offers promising preliminary results toward unsupervised detection of mental stress, we recognize a number of shortcomings that must be addressed with additional experiments and data. For instance, our training dataset was relatively small and additional data, including new modalities (e.g. motion, voice, etc.), would improve the accuracy of our trained models. Moreover, our method did not take into account intrinsic differences in HRV of different individuals which could be investigated with further experiments and data. In addition, it is imperative that the observed results in this work are thoroughly validated via new experiments.

## ACKNOWLEDGMENT

The authors would like to thank Maria Gabrani for her continuous support and useful discussions. We acknowledge support from IBM Research Frontiers Institute. The project leading to this publication has received funding from the European Union's Horizon 2020 research and innovation program under grant agreements no. 668858 and no. 826121.

## REFERENCES

[1] H.-G. Kim, E.-J. Cheon, D.-S. Bai, Y. H. Lee, and B.-H. Koo, "Stress and heart rate variability: A meta-analysis and review of the literature," *Psychiatry investigation*, vol. 15, no. 3, p. 235, 2018.

[2] J. Taelman, S. Vandeput, A. Spaepen, and S. Van Huffel, "Influence of mental stress on heart rate and heart rate variability," presented at the 4th European conference of the international federation for medical and biological engineering, 2009, pp. 1366–1369.

[3] F. Shaffer and J. P. Ginsberg, "An Overview of Heart Rate Variability Metrics and Norms," *Front Public Health*, vol. 5, pp. 258–258, Sep. 2017.

[4] U. Kraus *et al.*, "Individual daytime noise exposure during routine activities and heart rate variability in

adults: a repeated measures study," *Environmental health perspectives*, vol. 121, no. 5, pp. 607–612, 2013.

[5] W. von Rosenberg, T. Chanwimalueang, T. Adjei, U. Jaffer, V. Goverdovsky, and D. P. Mandic, "Resolving Ambiguities in the LF/HF Ratio: LF-HF Scatter Plots for the Categorization of Mental and Physical Stress from HRV," *Front Physiol*, vol. 8, pp. 360–360, Jun. 2017.

[6] M. Weippert, M. Kumar, S. Kreuzfeld, D. Arndt, A. Rieger, and R. Stoll, "Comparison of three mobile devices for measuring R–R intervals and heart rate variability: Polar S810i, Suunto t6 and an ambulatory ECG system," *European journal of applied physiology*, vol. 109, no. 4, pp. 779–786, 2010.

[7] R. Gevirtz, P. Lehrer, and M. Schwartz, "Cardiorespiratory biofeedback," *Biofeedback: A practitioner's guide*, pp. 196–213, 2016.

[8] R. McCraty and F. Shaffer, "Heart rate variability: new perspectives on physiological mechanisms, assessment of self-regulatory capacity, and health risk," *Global advances in health and medicine*, vol. 4, no. 1, pp. 46–61, 2015.

[9] S. Järvelin-Pasanen, S. Sinikallio, and M. P. Tarvainen, "Heart rate variability and occupational stress-systematic review," *Ind Health*, vol. 56, no. 6, pp. 500–511, Nov. 2018.

[10] M. Kivimäki, P. Leino-Arjas, R. Luukkonen, H. Riihimäi, J. Vahtera, and J. Kirjonen, "Work stress and risk of cardiovascular mortality: prospective cohort study of industrial employees," *Bmj*, vol. 325, no. 7369, p. 857, 2002.

[11] S. Salminen, M. Kivimäki, M. Elovainio, and J. Vahtera, "Stress factors predicting injuries of hospital personnel," *American journal of industrial medicine*, vol. 44, no. 1, pp. 32–36, 2003.

[12] H. Soori, M. Rahimi, and H. Mohseni, "Occupational stress and work-related unintentional injuries among Iranian car manufacturing workers," 2008.

[13] M. B. Harris, M. Baloğlu, and J. R. Stacks, "Mental health of trauma-exposed firefighters and critical incident stress debriefing," *Journal of Loss &Trauma*, vol. 7, no. 3, pp. 223–238, 2002.

[14] R. Beaton, S. Murphy, K. Pike, and M. Jarrett, "Stress-symptom factors in firefighters and paramedics.," 1995.

[15] R. Bhardwaj, P. Natrajan, and V. Balasubramanian, "Study to Determine the Effectiveness of Deep Learning Classifiers for ECG Based Driver Fatigue Classification," presented at the 2018 IEEE 13th International Conference on Industrial and Information Systems (ICIIS), 2018, pp. 98–102.

[16] A. Saeed, T. Ozcelebi, J. Lukkien, J. van Erp, and S. Trajanovski, "Model adaptation and personalization for physiological stress detection," presented at the 2018 IEEE 5th International Conference on Data Science and Advanced Analytics (DSAA), 2018, pp. 209–216.

[17] L. Wang and X. Zhou, "Detection of congestive heart failure based on LSTM-based deep network via short-term RR intervals," *Sensors*, vol. 19, no. 7, p. 1502, 2019.

[18] S.-H. Song and D. K. Kim, "Development of a Stress Classification Model Using Deep Belief Networks for Stress Monitoring," *Healthcare informatics research*, vol. 23, no. 4, pp. 285–292, 2017.

[19] B. Hwang, J. You, T. Vaessen, I. Myin-Germeys, C. Park, and B.-T. Zhang, "Deep ECGNet: An optimal deep learning framework for monitoring mental stress using ultra short-term ECG signals," *TELEMEDICINE and e-HEALTH*, vol. 24, no. 10, pp. 753–772, 2018.

[20] E. Smets *et al.*, "Comparison of machine learning techniques for psychophysiological stress detection," presented at the International Symposium on Pervasive Computing Paradigms for Mental Health, 2015, pp. 13–22.

[21] A. Shaw, N. Simsiri, I. Deznaby, M. Fiterau, and T. Rahaman, "Personalized Student Stress Prediction with Deep Multitask Network," *arXiv preprint arXiv:1906.11356*, 2019.

[22] L. Wang, W. Zhou, Q. Chang, J. Chen, and X. Zhou, "Deep Ensemble Detection of Congestive Heart Failure using Short-term RR Intervals," *IEEE Access*, 2019.

[23] B. M. Asl, S. K. Setarehdan, and M. Mohebbi, "Support vector machine-based arrhythmia classification using reduced features of heart rate variability signal," *Artificial Intelligence in Medicine*, vol. 44, no. 1, pp. 51–64, Sep. 2008.

[24] A. Y. Hannun *et al.*, "Cardiologist-level arrhythmia detection and classification in ambulatory electrocardiograms using a deep neural network," *Nature medicine*, vol. 25, no. 1, p. 65, 2019.

[25] D. Huysmans *et al.*, "Unsupervised Learning for Mental Stress Detection," presented at the Proceedings of the 11th International Joint Conference on Biomedical Engineering Systems and Technologies, 2018, vol. 4, pp. 26–35.

[26] L. Medina, "Identification of stress states from ECG signals using unsupervised learning methods," presented at the Portuguese Conf. on Pattern Recognition-RecPad, 2009.

[27] L. Salahuddin, J. Cho, M. G. Jeong, and D. Kim, "Ultra Short Term Analysis of Heart Rate Variability for Monitoring Mental Stress in Mobile Settings," in *2007 29th Annual International Conference of the IEEE Engineering in Medicine and Biology Society*, 2007, pp. 4656–4659.

[28] G. Giannakakis, D. Grigoriadis, K. Giannakaki, O. Simantiraki, A. Roniotis, and M. Tsiknakis, "Review on psychological stress detection using biosignals," *IEEE Transactions on Affective Computing*, 2019.

[29] J. C. C. Blásquez, G. R. Font, and L. C. Ortís, "Heart-rate variability and precompetitive anxiety in swimmers," *Psicothema*, vol. 21, no. 4, pp. 531–536, 2009.

[30] F.-T. Sun, C. Kuo, H.-T. Cheng, S. Buthpitiya, P. Collins, and M. Griss, “Activity-Aware Mental Stress Detection Using Physiological Sensors,” in *Mobile Computing, Applications, and Services*, 2012, pp. 282–301.

[31] C. C. Aggarwal, A. Hinneburg, and D. A. Keim, “On the surprising behavior of distance metrics in high dimensional space,” presented at the International conference on database theory, 2001, pp. 420–434.

[32] Y. P. Raykov, A. Boukouvalas, F. Baig, and M. A. Little, “What to Do When K-Means Clustering Fails: A Simple yet Principled Alternative Algorithm,” *PLOS ONE*, vol. 11, no. 9, p. e0162259, Sep. 2016.

[33] A. J. Camm *et al.*, “Heart rate variability: standards of measurement, physiological interpretation and clinical use. Task Force of the European Society of Cardiology and the North American Society of Pacing and Electrophysiology,” 1996.

[34] C. S. Weber *et al.*, “Low vagal tone is associated with impaired post stress recovery of cardiovascular, endocrine, and immune markers,” *European journal of applied physiology*, vol. 109, no. 2, pp. 201–211, 2010.

[35] S. Laborde, E. Mosley, and J. F. Thayer, “Heart Rate Variability and Cardiac Vagal Tone in Psychophysiological Research - Recommendations for Experiment Planning, Data Analysis, and Data Reporting,” *Front Psychol*, vol. 8, pp. 213–213, Feb. 2017.

[36] T. Kanungo, D. M. Mount, N. S. Netanyahu, C. D. Piatko, R. Silverman, and A. Y. Wu, “An efficient k-means clustering algorithm: Analysis and implementation,” *IEEE Transactions on Pattern Analysis & Machine Intelligence*, no. 7, pp. 881–892, 2002.

[37] F. Pedregosa *et al.*, “Scikit-learn: Machine learning in Python,” *Journal of machine learning research*, vol. 12, no. Oct, pp. 2825–2830, 2011.

[38] I. Goodfellow, Y. Bengio, and A. Courville, *Deep learning*. MIT press, 2016.

[39] D.-A. Clevert, T. Unterthiner, and S. Hochreiter, “Fast and accurate deep network learning by exponential linear units (elus),” *arXiv preprint arXiv:1511.07289*, 2015.

[40] E. Schubert, J. Sander, M. Ester, H. P. Kriegel, and X. Xu, “DBSCAN revisited, revisited: why and how you should (still) use DBSCAN,” *ACM Transactions on Database Systems (TODS)*, vol. 42, no. 3, p. 19, 2017.

[41] D. P. Kingma and J. Ba, “Adam: A method for stochastic optimization,” *arXiv preprint arXiv:1412.6980*, 2014.

# DeStress: Deep Learning for Unsupervised Identification of Mental Stress in Firefighters from Heart Rate Variability (HRV) Data

Ali Oskooei[1], Sophie Mai Chau [1], Jonas Weiss[1], Arvind Sridhar[1], María Rodríguez Martínez[1] and Bruno Michel[1]

[1] IBM Research – Zurich, Säumerstrasse 4, 8803 Rüschlikon, Switzerland

**Table S1: Time and frequency domain HRV features**

| Feature | Description |
|---|---|
| HFms | Absolute power of the high frequency band (0.15-0.4 Hz) |
| HFnu | Absolute power of the high frequency band (0.15-0.4 Hz) in normal units |
| HFpeak | Peak frequency of the high frequency band (0.15-0.4 Hz) |
| HFrel | Relative power of the high frequency band (0.15-0.4 Hz) |
| LF_HF | Ratio of LF to HF power |
| LFms | Absolute power of the low frequency band (0.0.04-0.15 Hz) |
| LFnu | Absolute power of the low frequency band (0.0.04-0.15 Hz) in normal units |
| LFpeak | Peak frequency of the low frequency band (0.0.04-0.15 Hz) |
| LFrel | Relative power of the low frequency band (0.0.04-0.15 Hz) |
| MaxRR | Maximum RR interval |
| MeanHR | Mean of successive heart rates |
| MeanRR | Mean of successive RR intervals |
| MinRR | Minimum RR interval |
| NN50 | Number of successive NN intervals that differ by more than 50 ms |
| RMSSD | The square root of the mean of the squares of the differences between successive RR intervals |
| SDNN | Mean of standard deviation for all normal RR intervals |
| VLFms | Absolute power of the very low frequency band (0.003 – 0.04 Hz) |
| pNN50 | Percentage of successive NN intervals that differ more than 50ms |